\documentclass[sigconf]{acmart}

\AtBeginDocument{%
  }

\usepackage{subfigure}

\copyrightyear{2024}
\acmYear{2024}
\setcopyright{acmlicensed}
\acmConference[CIKM '24]{Proceedings of the 33rd ACM International Conference on Information and Knowledge Management}{October 21--25, 2024}{Boise, ID, USA}
\acmBooktitle{Proceedings of the 33rd ACM International Conference on Information and Knowledge Management (CIKM '24), October 21--25, 2024, Boise, ID, USA}
\acmDOI{10.1145/3627673.3679992}
\acmISBN{979-8-4007-0436-9/24/10}

\begin{document}

\title{Scalable and Adaptive Spectral Embedding for\\ Attributed Graph Clustering}


\author{Yunhui Liu}
\author{Tieke He}
\authornote{Tieke He and Qing Wu are the corresponding authors.}
\affiliation{%
\institution{State Key Laboratory for Novel Software Technology}
\institution{Nanjing University}
\city{Nanjing}
\country{China}}
\email{hetieke@gmail.com}

\author{Qing Wu}
\authornotemark[1]
\affiliation{%
\institution{School of Business}
\institution{Nanjing University}
\city{Nanjing}
\country{China}}
\email{wuqing@nju.edu.cn}

\author{Tao Zheng}
\author{Jianhua Zhao}
\affiliation{%
\institution{State Key Laboratory for Novel Software Technology}
\institution{Nanjing University}
\city{Nanjing}
\country{China}}





\begin{abstract}
Attributed graph clustering, which aims to group the nodes of an attributed graph into disjoint clusters, has made promising advancements in recent years. However, most existing methods face challenges when applied to large graphs due to the expensive computational cost and high memory usage. In this paper, we introduce Scalable and Adaptive Spectral Embedding (SASE), a simple attributed graph clustering method devoid of parameter learning. SASE comprises three main components: node features smoothing via $k$-order simple graph convolution, scalable spectral clustering using random Fourier features, and adaptive order selection. With these designs, SASE not only effectively captures global cluster structures but also exhibits linear time and space complexity relative to the graph size. Empirical results demonstrate the superiority of SASE. For example, on the ArXiv dataset with 169K nodes and 1.17M edges, SASE achieves a 6.9\% improvement in ACC and a $5.87\times$ speedup compared to the runner-up, S3GC.
\end{abstract}

\ccsdesc[500]{Mathematics of computing~Graph algorithms}
\ccsdesc[500]{Information systems~Clustering}

\keywords{Attributed Graph Clustering; Spectral Embedding; Graph Neural Networks}


\maketitle

\section{Introduction}
Attributed graphs, comprising nodes associated with feature attributes and edges characterizing pairwise relationships, are natural and efficient representations of non-Euclidean data. Attributed graph clustering, which aims to partition nodes in such graphs into disjoint groups, has shown significant advancements in recent years \cite{DGCSurvey}. 

State-of-the-art attributed graph clustering methods focus on simultaneously modeling graph structure and node attributes to identify clusters within the data. In general, these methods can be classified into two categories: neural network (NN)-based and graph filtering (GF)-based. NN-based methods, such as graph auto-encoders \cite{GAE, DAEGC, RGAE} and contrastive graph clustering models \cite{GRACE, S3GC, CCGC}, first embed nodes into a latent space using graph neural networks \cite{GCN, SAGE, GAT} and then perform clustering algorithms on learned embeddings. Although proven effective, they rely on a large number of trainable parameters, suffering from expensive computation overhead and high memory usage. 

Instead, GF-based methods like AGC \cite{AGC}, FGC \cite{FGC}, and IAGC \cite{IAGC}, discard parameter learning. They first use a $k$-order graph convolution, which serves as a low-pass graph filter, to aggregate low-frequency signals from neighbors, thus obtaining smoothed node features. Subsequently, spectral clustering (SC) \cite{SC} is applied to the similarity graph constructed by the inner product of these smoothed features. However, these methods are constrained by two significant limitations: 1) Smoothed node features may neglect distinguishable information in the original node features, leading to suboptimal performance on graphs with more informative original features. 2) They exhibit poor scalability when applied to large-scale graphs, due to the SC's quadratic space and time complexity in constructing similarity graphs and computing subsequent eigendecomposition.

To tackle these issues, we propose Scalable and Adaptive Spectral Embedding (SASE) for attributed graph clustering. Specifically, we first conduct $k$-order simple graph convolution \cite{SGC} to smooth node features and fuse the original and smoothed node features linearly. Subsequently, we employ random Fourier features \cite{RFF} to explicitly project fused node features into the kernel space, where SC can be implicitly and efficiently performed without explicitly constructing similarity graphs and computing eigendecomposition. Furthermore, we take the mean ratio of the intra-cluster distance with respect to the nearest-cluster distance for each node as the selection criterion to adaptively determine the order $k$ of graph convolution. With the above designs, SASE not only effectively captures global cluster structures but also exhibits linear time and space complexity relative to the graph size. Empirical results demonstrate the superiority of SASE. For example, SASE achieves a 6.9\% ACC improvement and a 5.87\% speedup over the runner-up S3GC \cite{S3GC} on the ArXiv dataset with 169K nodes and 1.17M edges.

\section{Preliminaries}
\subsection{Problem Definition}
Consider an undirected attributed graph $\mathcal{G} = (\mathcal{V}, \mathcal{E}, \boldsymbol{X})$, where $\mathcal{V}=\{ v_i \}_{i=1}^{n}$ is a set of $n$ nodes, $\mathcal{E} \subset \mathcal{V} \times \mathcal{V}$ is a set of edges, and $\boldsymbol{X} \in \mathbb{R}^{n \times f}$ is the node attribute matrix. The adjacency matrix of the graph is denoted as $\boldsymbol{A} \in \{0,1\}^{n \times n}$, where $\boldsymbol{A}_{ij} = 1$ if an edge exists between node $v_i$ and $v_j$, and $\boldsymbol{A}_{ij} = 0$ otherwise. We take $\boldsymbol{I}$ to be an identity matrix, and $\boldsymbol{1}$ to be a column vector of ones. Our goal is to partition the node set $\mathcal{V}$ into $m$ disjoint groups $\mathcal{C} = \{ C_i \}_{i=1}^{m} = \{c_i\}_{i=1}^n$.

\subsection{Spectral Clustering} \label{Sec: SC}
Spectral clustering (SC) \cite{SC} is one of the most effective clustering approaches that capture hidden cluster structures in the data. It comprises two key stages: spectral embedding and subsequent k-means clustering, while spectral embedding involves the construction of a similarity graph and the eigendecomposition of the resulting graph Laplacian matrix. Given a dataset of $n$ data points $\boldsymbol{X} \in \mathbb{R}^{n \times f}$, SC constructs a fully connected graph $\boldsymbol{W} \in \mathbb{R}^{n \times n}$ using a similarity (kernel) function, e.g., the Gaussian kernel:
\begin{equation}
    \boldsymbol{W}_{ij} = \boldsymbol{\kappa} (\boldsymbol{x}_i,\boldsymbol{x}_j; \sigma)  = \exp \left( - \frac{|| \boldsymbol{x}_i - \boldsymbol{x}_j ||^2}{2\sigma^2} \right),
\end{equation}
where $\boldsymbol{\kappa}$ is the kernel function and $\sigma$ is the bandwidth parameter. Let $\boldsymbol{D} = \operatorname{diag}(\boldsymbol{W}\boldsymbol{1})$ be the diagonal degree matrix, and $\boldsymbol{L} = \boldsymbol{I} - \boldsymbol{D}^{-1/2} \boldsymbol{W} \boldsymbol{D}^{-1/2}$ be the normalized graph Laplacian matrix. Let $\boldsymbol{U} \in \mathbb{R}^{n \times m}$ denote the bottom $m$ eigenvectors of $\boldsymbol{L}$, or equivalently, the top $m$ eigenvectors of the
normalized adjacency matrix $\boldsymbol{D}^{-1/2} \boldsymbol{W} \boldsymbol{D}^{-1/2}$. SC groups the data points by performing k-means clustering on the spectral embedding matrix $\boldsymbol{\hat{U}}$, obtained from $\boldsymbol{U}$ through $\ell_2$ normalization. However, the high computational costs of the similarity matrix $\mathcal{O}(n^2f)$ and the eigendecomposition $\mathcal{O}(n^2m)$ make SC hardly scalable to large-scale datasets.

\subsection{Attributed Graph Clustering}
\textbf{Neural network-based methods} first train graph neural networks \cite{GCN, SAGE, GAT} using a label-agnostic loss to obtain low-dimensional node embeddings, followed by traditional or neural clustering on the learned embeddings. For instance, graph auto-encoders \cite{GAE, DAEGC, RGAE} learn node embeddings by reconstructing the graph adjacency matrix. However, the quadratic space and time complexity of the adjacency reconstruction loss limits their scalability. Contrastive graph clustering \cite{GRACE} learns node embeddings by aligning augmented node pairs (positive pairs) from the same node while separating any two distinct nodes (negative pairs) from the graph. CCGC \cite{CCGC} leverages intrinsic supervision information from high-confidence clustering results to enhance the quality of positive and negative pairs. To improve scalability, BGRL \cite{BGRL} eliminates the need for negative pairs by minimizing an invariance-based loss for augmented graphs within a batch, while S3GC \cite{S3GC} carefully designs a negative sampler and loss function for contrastive graph clustering. Despite their proven effectiveness, these NN-based methods rely on a large number of trainable parameters, leading to long training time and high memory usage.

\noindent \textbf{Graph filtering-based methods} first use a $k$-order low-pass graph filter to smooth node features and then apply traditional clustering algorithms such as k-means and SC \cite{SC} on the smoothed node features. For example, SGC \cite{SGC} utilizes the normalized adjacency matrix to aggregate long-range neighborhood information, while AGC \cite{AGC} introduces an adaptive low-pass graph filter. FGC \cite{AGC} further exploits high-order graph adjacency to learn a fine-grained self-expressive matrix. IAGC \cite{IAGC} boosts AGC by addressing the inconsistencies of filtered features with graph structure and raw features. However, these methods suffer from overlooking critical information in the original features and exhibit poor scalability on large graphs when employing SC. To address these issues, we propose a more effective and scalable GF-based method.

\section{Method}
\subsection{Node Features Smoothing} \label{Sec: NFS}
Classical graph neural networks \cite{GCN, SAGE, GAT} assume that connected nodes tend to have similar features and belong to the same cluster, thus node features are supposed to be smooth on the graph manifold. To this end, we can obtain clustering-friendly node features by smoothing original node features. For simplicity, we adopt a $k$-order simple graph convolution (SGC) \cite{SGC} as the low-pass graph filter to smooth node features:
\begin{equation}
    \boldsymbol{X}^{(k)} = \left(  \boldsymbol{\hat{D}}^{-1/2} \boldsymbol{\hat{A}}
        \boldsymbol{\hat{D}}^{-1/2} \right)^k \boldsymbol{X},
\end{equation}
where $\boldsymbol{\hat{A}}=\boldsymbol{A}+\boldsymbol{I}$ is the adjacency matrix with inserted self-loops, $\boldsymbol{\hat{D}}=\operatorname{diag}(\boldsymbol{\hat{A}}\boldsymbol{1})$ is the diagonal degree matrix, and $\boldsymbol{X}^{(k)} \in \mathbb{R}^{n \times f}$ denotes the smoothed feature matrix. SGC just discards trainable parameters in each layer of GCN \cite{GCN}, so it is faster and more scalable. Additionally, SGC aggregates long-range neighborhood information by applying the $k$-th power of the normalized adjacency matrix, which is proven to capture global cluster structures better \cite{AGC}.

However, we argue that smoothed node features may neglect key information in the original node features as the order $k$ increases, thus leading to suboptimal performance on graphs with more informative original features. To address this problem, we propose to combine the original and smoothed node features linearly:
\begin{equation}
    \boldsymbol{\widetilde{X}}  = \alpha \boldsymbol{X} + (1-\alpha) \boldsymbol{X}^{(k)},
\end{equation}
where $\alpha \in [0, 1]$ is the linear coefficient. Note that the resultant feature matrix $\boldsymbol{\widetilde{X}} \in \mathbb{R}^{n \times f}$ usually has a high dimensionality $f$, which will lead to the "curse of dimensionality" problem and additional computational cost in the following spectral clustering step. Therefore, we generate the final node features $\boldsymbol{Z} \in \mathbb{R}^{n \times d}$ by performing randomized truncated singular value decomposition (T-SVD) \cite{T-SVD} on $\boldsymbol{\widetilde{X}}$, where $d$ is the reduced dimensionality.

\subsection{Scalable Spectral Clustering} \label{Sec: SSC}
GF-based methods AGC \cite{AGC}, FGC \cite{FGC}, and IAGC \cite{IAGC} employ traditional SC directly on smoothed node features. However, SC encounters two significant computational bottlenecks: similarity graph construction and eigendecomposition, as discussed in Section \ref{Sec: SC}. Here, we aim to make SC scalable without explicitly constructing similarity graphs and computing eigendecomposition. 


Specifically, we first seek for a projection function $\boldsymbol{\phi}(\boldsymbol{z}): \mathbb{R}^d \to \mathbb{R}^{2D}$ such that the non-negative Gaussian kernel $\boldsymbol{\kappa}(\boldsymbol{z}_i, \boldsymbol{z}_j; \sigma) \approx \boldsymbol{\phi}(\boldsymbol{z}_i)^\top \boldsymbol{\phi}(\boldsymbol{z}_j)$. In other words, we can approximate the similarity graph adjacency $\boldsymbol{W}$ with $\boldsymbol{\widetilde{Z}} \boldsymbol{\widetilde{Z}}^\top$, where $\boldsymbol{\widetilde{Z}} = \{\boldsymbol{\phi}(\boldsymbol{z_i})\}_{i=1}^n$. Then we can efficiently perform SC using $\boldsymbol{\widetilde{Z}}$ by employing the following steps: 1) Compute the diagonal degree matrix: $\boldsymbol{\widetilde{D}} = \operatorname{diag}\left(\boldsymbol{\widetilde{Z}} \boldsymbol{\widetilde{Z}}^\top \boldsymbol{1}\right) = \operatorname{diag}\left(\boldsymbol{\widetilde{Z}} \left(\boldsymbol{\widetilde{Z}}^\top \boldsymbol{1}\right)\right)$; 2) Normalize $\boldsymbol{\widetilde{Z}}$ using the degree matrix $\boldsymbol{\widetilde{D}}$: $\boldsymbol{\hat{Z}} = \boldsymbol{\widetilde{D}}^{-1/2}\boldsymbol{\widetilde{Z}}$; 3) Compute the top $d$ left singular vectors $\boldsymbol{U}$ of $\boldsymbol{\hat{Z}}$,  which are equivalent to the bottom $d$ eigenvectors of the normalized graph Laplacian matrix $\boldsymbol{I}-\boldsymbol{\widetilde{D}}^{-1/2} \boldsymbol{\widetilde{Z}} \boldsymbol{\widetilde{Z}}^\top \boldsymbol{\widetilde{D}}^{-1/2}$; 4) Perform k-means on the $\ell_2$ normalized rows $\boldsymbol{\hat{U}}$ of $\boldsymbol{U}$. With these steps, SC can be performed with linear complexity w.r.t. the number of nodes, as detailed in Section \ref{Sec: Complexity Analysis}.

Then we discuss how to approximate the projection function $\boldsymbol{\phi}(\boldsymbol{z}): \mathbb{R}^d \to \mathbb{R}^{2D}$ through Random Fourier Features (RFF) \cite{RFF}. 
For the non-negative Gaussian kernel $\boldsymbol{\kappa}(\boldsymbol{z}_i, \boldsymbol{z}_j; \sigma)$, we first randomly sample $\{ \boldsymbol{\omega}_1, \boldsymbol{\omega}_2, \dots, \boldsymbol{\omega}_D \}$ from a Gaussian distribution $\mathcal{N}\left( \boldsymbol{0}, \frac{1}{\sigma}\boldsymbol{I} \right)$. Subsequently, the projection function is defined as:
\begin{equation}
    \boldsymbol{\phi}(\boldsymbol{z})=\frac{\left[ \cos(\boldsymbol{\omega}_1^\top \boldsymbol{z}), \dots, \cos(\boldsymbol{\omega}_D^\top \boldsymbol{z}),
    \sin(\boldsymbol{\omega}_1^\top \boldsymbol{z}), \dots, 
    \sin(\boldsymbol{\omega}_D^\top \boldsymbol{z})
    \right]^\top}{\sqrt{D}}.
\end{equation}
This approach explicitly projects node features into a kernel space, where SC can be efficiently performed using the resulting $\{\boldsymbol{\phi}(\boldsymbol{z_i})\}_{i=1}^n$. Additionally, alternative non-negative kernels like the quadratic and Laplacian kernels can be employed, or other kernel approximation techniques such as Nystroem Approximation \cite{Nystroem} and Tensor Sketch \cite{TensorSketch} can be considered.

\subsection{Adaptive Order Selection} \label{Sec: AOS}
The selection of the convolution order $k$ plays a pivotal role in our model. A small order may result in insufficient propagation of neighborhood information, while a large order can lead to over-smoothing, where the features of nodes from different clusters blend together and become indistinguishable \cite{AGC}. To address this concern, AGC proposes using the intra-cluster distance as the selection criterion to adaptively determine the order. However, this criterion overlooks inter-cluster structures and involves quadratic time complexity in computation. Therefore, we introduce a more comprehensive and scalable criterion. Given spectral embeddings $\{\boldsymbol{\hat{u}}_i\}_{i=1}^n$, k-means clustering assignments $\mathcal{C}=\{c_i\}_{i=1}^n$, and cluster centroids $\{ \boldsymbol{p}_i \}_{i=1}^m$, our criterion score is defined as:
\begin{equation}
    s=
    \frac{1}{n} \sum_{i=1}^n \frac{a(i)}{b(i)},
\end{equation}
where $a(i) = || \boldsymbol{\hat{u}}_i - \boldsymbol{p}_{c_i} ||$ is the distance of node $v_i$ to its cluster centroid $\boldsymbol{p}_{c_i}$, and $b(i) = \min_{c_i \not= j}||  \boldsymbol{\hat{u}}_i - \boldsymbol{p}_{j} ||$ is the distance of node $v_i$ to the centroid of its nearest cluster. $s \in [0,1]$ effectively captures global cluster structures since lower intra-cluster distance and higher inter-cluster distance will yield a better (smaller) score. Additionally, computing $s$ entails a time complexity of $\mathcal{O}(nmd)$, which scales linearly with the number of nodes.

Following AGC, we adaptively select the order $k$ as follows. Starting from $k = 1$, we increment it by $1$ iteratively. At each iteration $t$, we first obtain the clustering assignments $\mathcal{C}^{(t)}$, as outlined in Sections \ref{Sec: NFS} and \ref{Sec: SSC}. Subsequently, we compute the criterion score $s^{(t)}$. Once $\Delta = s^{(t-1)}-s^{(t)} < 0$, we stop the iteration and set the chosen $k=t-1$. Consequently, the final clustering result is $\mathcal{C}^{(t-1)}$.

\subsection{Complexity Analysis} \label{Sec: Complexity Analysis}
\noindent\textbf{Space Complexity.} \quad
The space complexity for SASE is $\mathcal{O}(|\mathcal{E}|+n(f+d+D))$, where $|\mathcal{E}|$ is the total edge count, $nf$ accounts for storing fused node features $\boldsymbol{\widetilde{X}}$, $nd$ for node embeddings $\boldsymbol{Z}$ and spectral embeddings $\boldsymbol{\hat{U}}$, and $nD$ for projected node embeddings $\boldsymbol{\widetilde{Z}}$.

\noindent\textbf{Time Complexity.} \quad  
The time complexity is $\mathcal{O}( k|\mathcal{E}|f + nkf + n(f+D)\log(d)+n(m+D)d)$. Specifically, computing mixed node features $\boldsymbol{\widetilde{X}} \in \mathbb{R}^{n \times f}$ takes $\mathcal{O}(k|\mathcal{E}|f+knf)$ time; performing randomized T-SVD \cite{T-SVD} on $\boldsymbol{\widetilde{X}}$ to obtain $\boldsymbol{Z} \in \mathbb{R}^{n \times d}$ takes $\mathcal{O}(nf\log(d))$ time; projecting $\boldsymbol{Z}$ to $\boldsymbol{\widetilde{Z}} \in \mathbb{R}^{n \times 2D}$ using RFF takes $\mathcal{O}(ndD)$ time; computing the diagonal degree matrix $\boldsymbol{\widetilde{D}}$ and $\boldsymbol{\hat{Z}}$ takes $\mathcal{O}(nD)$ time; randomized T-SVD on $\boldsymbol{\hat{Z}}$ for $\boldsymbol{U} \in \mathbb{R}^{n \times d}$ takes $\mathcal{O}(nD\log(d))$ time; applying k-means clustering to $\boldsymbol{\hat{U}}$ roughly takes $\mathcal{O}(nmd)$ time; computing the selection criterion $s$ takes $\mathcal{O}(nmd)$ time.

In summary, SASE can be executed with linear space and time complexity w.r.t. the graph size. Additionally, SASE does not have any trainable parameters. Therefore, SASE enjoys very high efficiency and scalability.

\begin{table*}[!ht]
    \begin{center}
    \setlength{\tabcolsep}{3.5pt}
     \renewcommand\arraystretch{0.9}
    {\caption{Overall performance on node clustering (in percentage). "-" denotes that the method ran Out of Memory.}\label{Tab: Node clustering}}
    \vspace{-0.1cm}
    \begin{tabular}{c|ccc|ccc|ccc|ccc}
    \bottomrule
    Dataset     & \multicolumn{3}{c|}{Cora}      & \multicolumn{3}{c|}{CiteSeer}        & \multicolumn{3}{c|}{PubMed}      & \multicolumn{3}{c}{ArXiv} \\ \hline
    Metric      & ACC    & NMI    & ARI          & ACC    & NMI    & ARI                & ACC    & NMI    & ARI            & ACC    & NMI    & ARI      \\ \hline
    k-means     & 34.1±3.2 & 15.4±3.8 & 9.5±2.0  & 43.8±3.6 & 20.7±2.8 & 16.8±3.0 & 60.1±0.0 & 31.3±0.2 & 28.1±0.0 & 17.6±0.4 & 21.6±0.4 & 7.4±0.5   \\
    SC          & 59.1±1.2 & 46.1±1.0 & 34.4±1.0 & 46.0±0.7 & 23.9±0.5 & 18.5±0.4 & 66.6±0.0 & 28.8±0.0 & 30.3±0.0 & - & - & - \\
    DMoN        & 51.7±2.9 & 47.3±1.0 & 30.1±1.2 & 38.5±1.7 & 30.3±2.7 & 20.0±2.8 & 35.1±1.8 & 25.7±1.9 & 10.8±1.2 & 25.0±0.7 & 35.6±0.8 & 12.7±0.3 \\
    SENet       & 71.9±0.7 & 55.1±0.7 & 49.0±1.1 & 67.5±0.8 & 41.7±0.8 & 42.4±1.0 & 67.6±0.6 & 30.6±1.5 & 29.7±1.1 & - & - & - \\
    GAE         & 64.5±1.9 & 52.4±1.3 & 43.9±2.3 & 59.9±0.7 & 34.5±0.9 & 32.7±0.9 & 66.1±0.7 & 27.2±1.5 & 26.3±1.2 & - & - & - \\
    DAEGC       & 70.4±0.4 & 52.9±0.7 & 45.6±0.4 & 67.5±1.4 & 39.4±0.9 & 40.8±1.2 & 67.1±1.0 & 26.6±1.4 & 27.7±1.2 & - & - & - \\
    RARGA & 71.2±0.7 & 50.7±0.8 & 47.1±2.3 & 48.6±0.7 & 28.5±0.3 & 18.9±1.3 & 69.2±0.9 & 30.0±1.2 & 30.9±1.4 & - & - & - \\
    GRACE       & 68.3±0.3 & 52.0±0.2 & 46.2±0.5 & 65.8±0.1 & 39.1±0.1 & 40.4±0.1 & 63.7±0.0 & 30.8±0.0 & 27.6±0.0 & - & - & - \\
    CCGC        & 73.9±1.2 & 56.5±1.0 & 52.5±1.9 & 69.8±0.9 & 44.3±0.8 & 45.7±1.8 & 68.8±0.6 & 32.1±0.2 & 30.8±0.4 & - & - & - \\
    BGRL        & 68.8±1.8 & 53.2±1.5 & 46.2±2.9 & 65.6±1.2 & 41.4±1.3 & 38.7±1.9 & 65.8±0.0 & 30.1±0.0 & 27.3±0.0 & 22.7±1.0 & 32.1±0.5 & 13.0±1.2 \\
    DGI & 70.9±1.0 & 55.8±0.6 & 48.9±1.4 & 68.6±1.0 & 43.5±0.9 & 44.5±1.6 & 65.7±1.0 & 32.2±0.8 & 29.2±2.0 & 31.4±1.9 & 41.2±0.8 & 22.3±2.7 \\
    S3GC        & \textbf{74.2±0.8} & \textbf{58.8±0.9} & \textbf{54.4±1.2} & 68.8±1.0 & 44.1±1.1 & 44.8±0.9 & \textbf{71.3±0.9} & 33.3±1.0 & 34.5±0.9 & 35.0±1.2 & 46.3±0.3 & 27.0±1.6 \\ \hline
    SGC & 64.9±2.0 & 51.8±1.3 & 43.7±2.0 & 68.3±0.2 & 43.1±0.2 & 43.7±0.2 & 68.5±0.0 & 31.9±0.0 & 31.0±0.0 & 38.2±0.5 & 45.2±0.2 & 33.3±0.5 \\
    AGC         & 68.9±0.2 & 53.7±0.4 & 44.8±0.0 & 67.0±0.2 & 41.1±0.4 & 41.6±0.0 & 69.8±0.0 & 31.6±0.0 & 31.0±0.0 & - & - & - \\
    FGC         & 69.3±0.0 & 54.2±0.1 & 47.0±0.1 & 68.2±0.0 & 43.2±0.1 & 43.9±0.1 & 69.2±0.0 & 32.1±0.0 & 31.5±0.0 & - & - & - \\
    IAGC        & 72.4±0.0 & 55.7±0.0 & 49.3±0.0 & 69.1±0.0 & 43.1±0.0 & 44.3±0.0 & 70.5±0.0 & 31.6±0.0 & 33.0±0.0 & - & - & - \\ 
    SASE        & 71.4±1.9 & 55.9±1.2 & 48.7±1.7 & \textbf{70.2±0.1} & \textbf{44.9±0.1} & \textbf{46.8±0.1} & \textbf{71.3±0.0} & \textbf{37.0±0.0} & \textbf{35.3±0.0} & \textbf{41.9±0.6} & \textbf{46.8±0.1} & \textbf{37.6±0.4} \\
    \toprule
    \end{tabular}
    \vspace{-0.1cm}
    \end{center}
\end{table*}

\section{Experiments}
\subsection{Experimental Setup}

\subsubsection{Datasets}
We  conduct experiments on 3 small datasets: Cora, CiteSeer, PubMed \cite{GCN}, and 1 large dataset: ArXiv \cite{OGB}. They are citation networks with nodes as papers and edges as citations. All nodes are labeled based on the respective subjects.


\subsubsection{Baselines}
We compare SASE with two traditional clustering methods k-means and SC \cite{SC}; ten NN-based methods DMoN \cite{DMoN}, SENet \cite{SENet}, GAE \cite{GAE}, DAEGC \cite{DAEGC}, RARGA \cite{RGAE}, GRACE \cite{GRACE}, CCGC \cite{CCGC}, BGRL \cite{BGRL}, DGI \cite{DGI}, and S3GC \cite{S3GC}; four GF-based methods SGC \cite{SGC}, AGC \cite{AGC}, FGC \cite{FGC}, and IAGC \cite{IAGC}. Their results are either taken from the published paper or obtained using official codes when there are no corresponding results.

\subsubsection{Implementation Details}
We implement SASE using PyTorch on a 32GB NVIDIA Tesla V100 GPU, aligning cluster numbers with ground-truth classes and evaluating using Accuracy, NMI, and ARI. Hyper-parameters include setting $2D$ projection dimensionality to $100$ and a maximum iteration number of $50$, with others tuned via grid search. 


\begin{table}[h]
    \begin{center}
    {\caption{Dataset statistics and hyper-parameters.}\label{Tab: Dataset statistics}}
    \vspace{-0.1cm}
    \begin{tabular}{l|cccc|cccc}
    \bottomrule
    Dataset    & \#Nodes    & \#Edges   & \#Feat.  & \#Clus.  & $k$   & $\alpha$ & $d$\\
    \hline 
    Cora       & 2,708      & 5,278     & 1,433       & 7           & 12    & 0.20     & 32 \\
    CiteSeer   & 3,327      & 4,614     & 3,703       & 6           & 11    & 0.30     & 12 \\
    PubMed     & 19,717     & 44,325    & 500         & 3           & 43    & 0.30     & 3  \\
    ArXiv      & 169,343	& 1,166,243 & 128         & 40          & 11    & 0.05     & 64 \\
    \toprule
    \end{tabular}
    \vspace{-0.1cm}
    \end{center}
\end{table}

\subsection{Experimental Results}

\subsubsection{Performance}
As shown in Table \ref{Tab: Node clustering}, SASE surpasses all baselines on three datasets, except Cora. We attribute this exception to its requirement for powerful fitting abilities in non-linear neural networks. Compared to the runner-up S3GC, SASE improves NMI by 3.7\% on PubMed and ACC by 6.9\% on ArXiv, since SASE can aggregate long-range neighborhood information via $k$-order graph convolution while S3GC utilizes only a 1-layer GCN. Additionally, in contrast to GF-based methods, SASE achieves better performance on CiteSeer and PubMed. This superiority arises from SASE's capacity to retain crucial information in the original node features, as verified in Section \ref{Sec: Effectiveness of FOF}.

\subsubsection{Scalability}\label{Sec: Scalability}
In Table \ref{Tab: Execution Time}, we report the time costs of various methods on PubMed and ArXiv. It is observed that: 1) SASE greatly improves the scalability of GF-based methods AGC, FGC, and IAGC, which take quadratic space complexity during SC, e.g., storing the dense similarity graph for SC on ArXiv requires about 107 GiB of memory. 2) SASE demonstrates significant speed advantages over NN-based methods. For example, SASE achieves a $15.61\times$ speedup over DGI on PubMed and a $5.87\times$ speedup over S3GC on ArXiv.

\begin{table}[h]
    \begin{center}
    \setlength{\tabcolsep}{1pt}
    {\caption{Execution time of various methods in seconds.}\label{Tab: Execution Time}}
    \vspace{-0.1cm}
    \begin{tabular}{l|cccc|ccccc}
    \bottomrule
    Dataset    & DMoN & BGRL & DGI & S3GC & SGC & AGC & FGC & IAGC & SASE \\
    \hline 
    PubMed     & 33.60 & 376.85 & 6.40 & 28.84 & 1.92 & 26.72 & 90.82 & 22.99 & \textbf{0.41} \\
    ArXiv      & 1850.49 & - & 196.19 & 193.05 & 40.46 & - & - & - & \textbf{32.91}  \\
    \toprule
    \end{tabular}
    \vspace{-0.1cm}
    \end{center}
\end{table}

\subsubsection{Effectiveness of Fusing Original Features}\label{Sec: Effectiveness of FOF}
Figure \ref{Fig: Effectiveness}(a) shows the variation in NMI w.r.t. different weights $\alpha$ assigned to original node features. Initially, performance improves with increasing $\alpha$ but then declines. This observation highlights the efficacy of combining original and smoothed node features, particularly on datasets with highly informative node features. For example, SASE with $\alpha>0$ consistently outperforms SASE with $\alpha=0$ on PubMed.

\subsubsection{Effectiveness of Adaptive Order Selection}\label{Sec: Effectiveness of AOS}
\begin{figure}[h]
    \centering
    \subfigure[NMI w.r.t. $\alpha$ on CiteSeer and PubMed]{\includegraphics[width=0.325\linewidth]{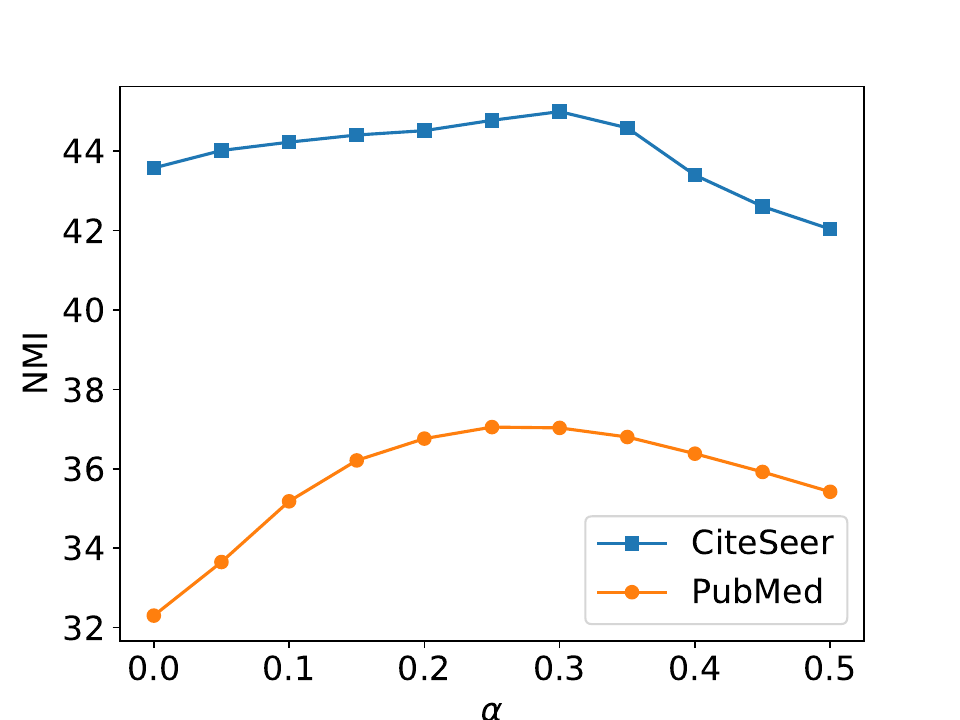}}
    \subfigure[NMI and $\Delta$ w.r.t. $k$ on CiteSeer]{\includegraphics[width=0.325\linewidth]{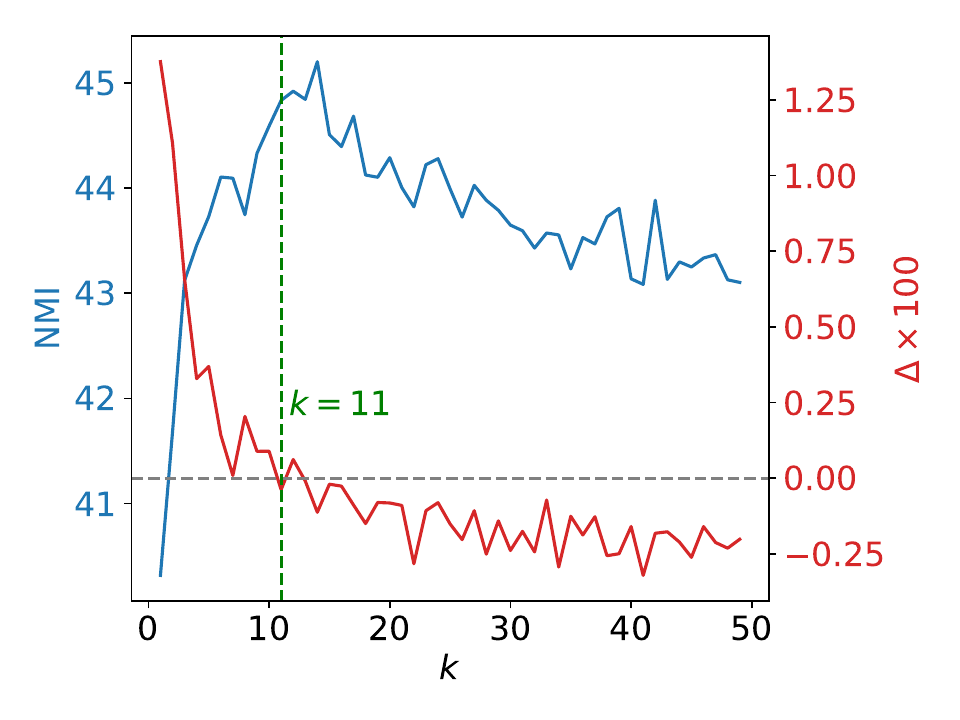}}
    \subfigure[NMI and $\Delta$ w.r.t. $k$ on PubMed]{\includegraphics[width=0.325\linewidth]{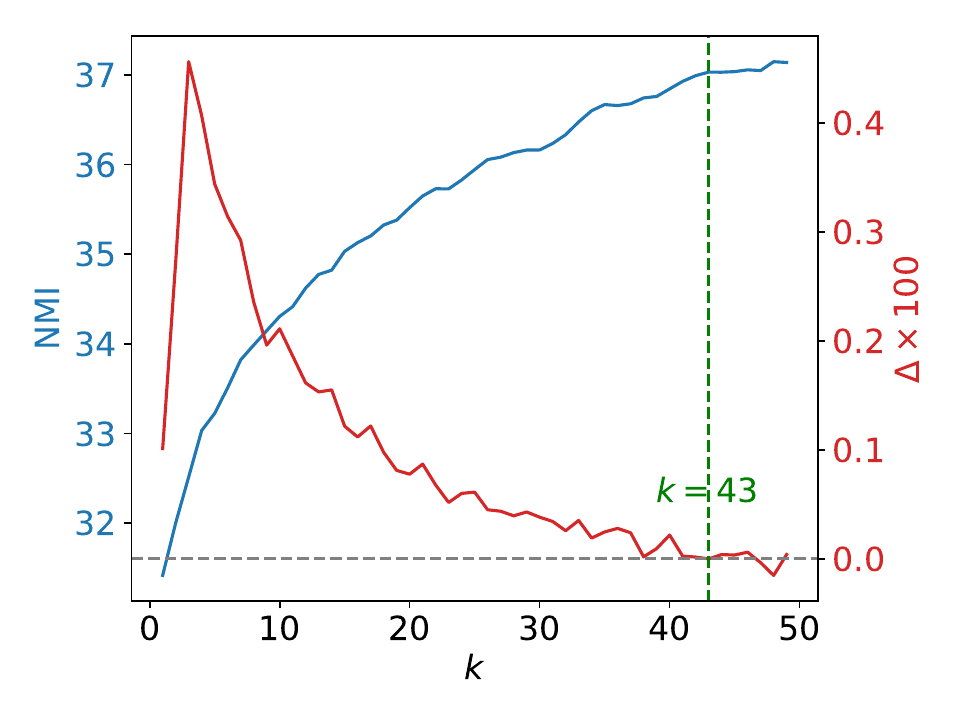}}
    \caption{Impact of $\alpha$ and $k$ on CiteSeer and PubMed.}\label{Fig: Effectiveness}
\end{figure}

Figures \ref{Fig: Effectiveness}(b) and (c) depict the variations in NMI and $\Delta$ w.r.t. different orders $k$. We can see that when $\Delta < 0$, the corresponding NMI values closely approach the best performance. The selected $k$ for Citeseer and PubMed, 11 and 43 respectively, approximate the optimal values of 13 and 48 on these datasets. These affirm the reliability of our proposed selection criterion in finding a favorable cluster partition.

\section{Conclusion}

This paper presents SASE, which comprises node features smoothing via $k$-order simple graph convolution, scalable spectral clustering using random Fourier features, and adaptive order selection. SASE not only effectively captures global cluster structures but also exhibits linear time and space complexity relative to the graph size. 
Empirical results demonstrate SASE's effectiveness and efficiency.

\begin{acks}
This work is partially supported by the National Key Research and Development Program of China (2021YFB1715600), and the National Natural Science Foundation of China (62306137).
\end{acks}


\bibliographystyle{ACM-Reference-Format}
\balance
\bibliography{references}


\begin{thebibliography}{24}


\ifx \showCODEN    \undefined \def \showCODEN     #1{\unskip}     \fi
\ifx \showDOI      \undefined \def \showDOI       #1{#1}\fi
\ifx \showISBNx    \undefined \def \showISBNx     #1{\unskip}     \fi
\ifx \showISBNxiii \undefined \def \showISBNxiii  #1{\unskip}     \fi
\ifx \showISSN     \undefined \def \showISSN      #1{\unskip}     \fi
\ifx \showLCCN     \undefined \def \showLCCN      #1{\unskip}     \fi
\ifx \shownote     \undefined \def \shownote      #1{#1}          \fi
\ifx \showarticletitle \undefined \def \showarticletitle #1{#1}   \fi
\ifx \showURL      \undefined \def \showURL       {\relax}        \fi
\providecommand\bibfield[2]{#2}
\providecommand\bibinfo[2]{#2}
\providecommand\natexlab[1]{#1}
\providecommand\showeprint[2][]{arXiv:#2}

\bibitem[Devvrit et~al\mbox{.}(2022)]%
        {S3GC}
\bibfield{author}{\bibinfo{person}{Fnu Devvrit}, \bibinfo{person}{Aditya Sinha}, \bibinfo{person}{Inderjit Dhillon}, {and} \bibinfo{person}{Prateek Jain}.} \bibinfo{year}{2022}\natexlab{}.
\newblock \showarticletitle{S3GC: scalable self-supervised graph clustering}.
\newblock \bibinfo{journal}{\emph{Advances in Neural Information Processing Systems}}  \bibinfo{volume}{35} (\bibinfo{year}{2022}), \bibinfo{pages}{3248--3261}.
\newblock


\bibitem[Halko et~al\mbox{.}(2011)]%
        {T-SVD}
\bibfield{author}{\bibinfo{person}{Nathan Halko}, \bibinfo{person}{Per-Gunnar Martinsson}, {and} \bibinfo{person}{Joel~A Tropp}.} \bibinfo{year}{2011}\natexlab{}.
\newblock \showarticletitle{Finding structure with randomness: Probabilistic algorithms for constructing approximate matrix decompositions}.
\newblock \bibinfo{journal}{\emph{SIAM review}} \bibinfo{volume}{53}, \bibinfo{number}{2} (\bibinfo{year}{2011}), \bibinfo{pages}{217--288}.
\newblock


\bibitem[Hamilton et~al\mbox{.}(2017)]%
        {SAGE}
\bibfield{author}{\bibinfo{person}{Will Hamilton}, \bibinfo{person}{Zhitao Ying}, {and} \bibinfo{person}{Jure Leskovec}.} \bibinfo{year}{2017}\natexlab{}.
\newblock \showarticletitle{Inductive representation learning on large graphs}.
\newblock \bibinfo{journal}{\emph{Advances in neural information processing systems}}  \bibinfo{volume}{30} (\bibinfo{year}{2017}).
\newblock


\bibitem[Hu et~al\mbox{.}(2020)]%
        {OGB}
\bibfield{author}{\bibinfo{person}{Weihua Hu}, \bibinfo{person}{Matthias Fey}, \bibinfo{person}{Marinka Zitnik}, \bibinfo{person}{Yuxiao Dong}, \bibinfo{person}{Hongyu Ren}, \bibinfo{person}{Bowen Liu}, \bibinfo{person}{Michele Catasta}, {and} \bibinfo{person}{Jure Leskovec}.} \bibinfo{year}{2020}\natexlab{}.
\newblock \showarticletitle{Open graph benchmark: Datasets for machine learning on graphs}.
\newblock \bibinfo{journal}{\emph{Advances in neural information processing systems}}  \bibinfo{volume}{33} (\bibinfo{year}{2020}), \bibinfo{pages}{22118--22133}.
\newblock


\bibitem[Kang et~al\mbox{.}(2022)]%
        {FGC}
\bibfield{author}{\bibinfo{person}{Zhao Kang}, \bibinfo{person}{Zhanyu Liu}, \bibinfo{person}{Shirui Pan}, {and} \bibinfo{person}{Ling Tian}.} \bibinfo{year}{2022}\natexlab{}.
\newblock \showarticletitle{Fine-grained attributed graph clustering}. In \bibinfo{booktitle}{\emph{Proceedings of the 2022 SIAM International Conference on Data Mining (SDM)}}. SIAM, \bibinfo{pages}{370--378}.
\newblock


\bibitem[Kipf and Welling(2016)]%
        {GAE}
\bibfield{author}{\bibinfo{person}{Thomas~N Kipf} {and} \bibinfo{person}{Max Welling}.} \bibinfo{year}{2016}\natexlab{}.
\newblock \showarticletitle{Variational graph auto-encoders}.
\newblock \bibinfo{journal}{\emph{arXiv preprint arXiv:1611.07308}} (\bibinfo{year}{2016}).
\newblock


\bibitem[Kipf and Welling(2017)]%
        {GCN}
\bibfield{author}{\bibinfo{person}{Thomas~N. Kipf} {and} \bibinfo{person}{Max Welling}.} \bibinfo{year}{2017}\natexlab{}.
\newblock \showarticletitle{Semi-Supervised Classification with Graph Convolutional Networks}. In \bibinfo{booktitle}{\emph{International Conference on Learning Representations}}.
\newblock


\bibitem[Liu et~al\mbox{.}(2022)]%
        {DGCSurvey}
\bibfield{author}{\bibinfo{person}{Yue Liu}, \bibinfo{person}{Jun Xia}, \bibinfo{person}{Sihang Zhou}, \bibinfo{person}{Xihong Yang}, \bibinfo{person}{Ke Liang}, \bibinfo{person}{Chenchen Fan}, \bibinfo{person}{Yan Zhuang}, \bibinfo{person}{Stan~Z Li}, \bibinfo{person}{Xinwang Liu}, {and} \bibinfo{person}{Kunlun He}.} \bibinfo{year}{2022}\natexlab{}.
\newblock \showarticletitle{A Survey of Deep Graph Clustering: Taxonomy, Challenge, Application, and Open Resource}.
\newblock \bibinfo{journal}{\emph{arXiv preprint arXiv:2211.12875}} (\bibinfo{year}{2022}).
\newblock


\bibitem[Mrabah et~al\mbox{.}(2022)]%
        {RGAE}
\bibfield{author}{\bibinfo{person}{Nairouz Mrabah}, \bibinfo{person}{Mohamed Bouguessa}, \bibinfo{person}{Mohamed~Fawzi Touati}, {and} \bibinfo{person}{Riadh Ksantini}.} \bibinfo{year}{2022}\natexlab{}.
\newblock \showarticletitle{Rethinking graph auto-encoder models for attributed graph clustering}.
\newblock \bibinfo{journal}{\emph{IEEE Transactions on Knowledge and Data Engineering}} (\bibinfo{year}{2022}).
\newblock


\bibitem[M{\"u}ller(2023)]%
        {DMoN}
\bibfield{author}{\bibinfo{person}{Emmanuel M{\"u}ller}.} \bibinfo{year}{2023}\natexlab{}.
\newblock \showarticletitle{Graph clustering with graph neural networks}.
\newblock \bibinfo{journal}{\emph{Journal of Machine Learning Research}}  \bibinfo{volume}{24} (\bibinfo{year}{2023}), \bibinfo{pages}{1--21}.
\newblock


\bibitem[Ng et~al\mbox{.}(2001)]%
        {SC}
\bibfield{author}{\bibinfo{person}{Andrew Ng}, \bibinfo{person}{Michael Jordan}, {and} \bibinfo{person}{Yair Weiss}.} \bibinfo{year}{2001}\natexlab{}.
\newblock \showarticletitle{On spectral clustering: Analysis and an algorithm}.
\newblock \bibinfo{journal}{\emph{Advances in neural information processing systems}}  \bibinfo{volume}{14} (\bibinfo{year}{2001}).
\newblock


\bibitem[Pham and Pagh(2013)]%
        {TensorSketch}
\bibfield{author}{\bibinfo{person}{Ninh Pham} {and} \bibinfo{person}{Rasmus Pagh}.} \bibinfo{year}{2013}\natexlab{}.
\newblock \showarticletitle{Fast and scalable polynomial kernels via explicit feature maps}. In \bibinfo{booktitle}{\emph{Proceedings of the 19th ACM SIGKDD international conference on Knowledge discovery and data mining}}. \bibinfo{pages}{239--247}.
\newblock


\bibitem[Rahimi and Recht(2007)]%
        {RFF}
\bibfield{author}{\bibinfo{person}{Ali Rahimi} {and} \bibinfo{person}{Benjamin Recht}.} \bibinfo{year}{2007}\natexlab{}.
\newblock \showarticletitle{Random features for large-scale kernel machines}.
\newblock \bibinfo{journal}{\emph{Advances in neural information processing systems}}  \bibinfo{volume}{20} (\bibinfo{year}{2007}).
\newblock


\bibitem[Thakoor et~al\mbox{.}(2022)]%
        {BGRL}
\bibfield{author}{\bibinfo{person}{Shantanu Thakoor}, \bibinfo{person}{Corentin Tallec}, \bibinfo{person}{Mohammad~Gheshlaghi Azar}, \bibinfo{person}{Mehdi Azabou}, \bibinfo{person}{Eva~L Dyer}, \bibinfo{person}{Remi Munos}, \bibinfo{person}{Petar Veli{\v{c}}kovi{\'c}}, {and} \bibinfo{person}{Michal Valko}.} \bibinfo{year}{2022}\natexlab{}.
\newblock \showarticletitle{Large-Scale Representation Learning on Graphs via Bootstrapping}. In \bibinfo{booktitle}{\emph{International Conference on Learning Representations}}.
\newblock


\bibitem[Veličković et~al\mbox{.}(2018)]%
        {GAT}
\bibfield{author}{\bibinfo{person}{Petar Veličković}, \bibinfo{person}{Guillem Cucurull}, \bibinfo{person}{Arantxa Casanova}, \bibinfo{person}{Adriana Romero}, \bibinfo{person}{Pietro Liò}, {and} \bibinfo{person}{Yoshua Bengio}.} \bibinfo{year}{2018}\natexlab{}.
\newblock \showarticletitle{Graph Attention Networks}. In \bibinfo{booktitle}{\emph{International Conference on Learning Representations}}.
\newblock


\bibitem[Veličković et~al\mbox{.}(2019)]%
        {DGI}
\bibfield{author}{\bibinfo{person}{Petar Veličković}, \bibinfo{person}{William Fedus}, \bibinfo{person}{William~L. Hamilton}, \bibinfo{person}{Pietro Liò}, \bibinfo{person}{Yoshua Bengio}, {and} \bibinfo{person}{R~Devon Hjelm}.} \bibinfo{year}{2019}\natexlab{}.
\newblock \showarticletitle{Deep Graph Infomax}. In \bibinfo{booktitle}{\emph{International Conference on Learning Representations}}.
\newblock
\urldef\tempurl%
\url{https://openreview.net/forum?id=rklz9iAcKQ}
\showURL{%
\tempurl}


\bibitem[Wang et~al\mbox{.}(2019)]%
        {DAEGC}
\bibfield{author}{\bibinfo{person}{Chun Wang}, \bibinfo{person}{Shirui Pan}, \bibinfo{person}{Ruiqi Hu}, \bibinfo{person}{Guodong Long}, \bibinfo{person}{Jing Jiang}, {and} \bibinfo{person}{Chengqi Zhang}.} \bibinfo{year}{2019}\natexlab{}.
\newblock \showarticletitle{Attributed Graph Clustering: A Deep Attentional Embedding Approach}. In \bibinfo{booktitle}{\emph{Proceedings of the 28th International Joint Conference on Artificial Intelligence}} \emph{(\bibinfo{series}{IJCAI'19})}. \bibinfo{pages}{3670–3676}.
\newblock


\bibitem[Williams and Seeger(2000)]%
        {Nystroem}
\bibfield{author}{\bibinfo{person}{Christopher Williams} {and} \bibinfo{person}{Matthias Seeger}.} \bibinfo{year}{2000}\natexlab{}.
\newblock \showarticletitle{Using the Nystr{\"o}m method to speed up kernel machines}.
\newblock \bibinfo{journal}{\emph{Advances in neural information processing systems}}  \bibinfo{volume}{13} (\bibinfo{year}{2000}).
\newblock


\bibitem[Wu et~al\mbox{.}(2019)]%
        {SGC}
\bibfield{author}{\bibinfo{person}{Felix Wu}, \bibinfo{person}{Amauri Souza}, \bibinfo{person}{Tianyi Zhang}, \bibinfo{person}{Christopher Fifty}, \bibinfo{person}{Tao Yu}, {and} \bibinfo{person}{Kilian Weinberger}.} \bibinfo{year}{2019}\natexlab{}.
\newblock \showarticletitle{Simplifying graph convolutional networks}. In \bibinfo{booktitle}{\emph{International conference on machine learning}}. PMLR, \bibinfo{pages}{6861--6871}.
\newblock


\bibitem[Yang et~al\mbox{.}(2023)]%
        {CCGC}
\bibfield{author}{\bibinfo{person}{Xihong Yang}, \bibinfo{person}{Yue Liu}, \bibinfo{person}{Sihang Zhou}, \bibinfo{person}{Siwei Wang}, \bibinfo{person}{Wenxuan Tu}, \bibinfo{person}{Qun Zheng}, \bibinfo{person}{Xinwang Liu}, \bibinfo{person}{Liming Fang}, {and} \bibinfo{person}{En Zhu}.} \bibinfo{year}{2023}\natexlab{}.
\newblock \showarticletitle{Cluster-guided Contrastive Graph Clustering Network}. In \bibinfo{booktitle}{\emph{AAAI Conference on Artificial Intelligence}}.
\newblock


\bibitem[Zhang et~al\mbox{.}(2019)]%
        {AGC}
\bibfield{author}{\bibinfo{person}{Xiaotong Zhang}, \bibinfo{person}{Han Liu}, \bibinfo{person}{Qimai Li}, {and} \bibinfo{person}{Xiao-Ming Wu}.} \bibinfo{year}{2019}\natexlab{}.
\newblock \showarticletitle{Attributed graph clustering via adaptive graph convolution}. In \bibinfo{booktitle}{\emph{Proceedings of the 28th International Joint Conference on Artificial Intelligence}} \emph{(\bibinfo{series}{IJCAI'19})}. \bibinfo{pages}{4327–4333}.
\newblock


\bibitem[Zhang et~al\mbox{.}(2023)]%
        {IAGC}
\bibfield{author}{\bibinfo{person}{Xiaotong Zhang}, \bibinfo{person}{Han Liu}, \bibinfo{person}{Qimai Li}, \bibinfo{person}{Xiao-Ming Wu}, {and} \bibinfo{person}{Xianchao Zhang}.} \bibinfo{year}{2023}\natexlab{}.
\newblock \showarticletitle{Adaptive Graph Convolution Methods for Attributed Graph Clustering}.
\newblock \bibinfo{journal}{\emph{IEEE Transactions on Knowledge and Data Engineering}} (\bibinfo{year}{2023}).
\newblock


\bibitem[Zhang et~al\mbox{.}(2021)]%
        {SENet}
\bibfield{author}{\bibinfo{person}{Xiaotong Zhang}, \bibinfo{person}{Han Liu}, \bibinfo{person}{Xiao-Ming Wu}, \bibinfo{person}{Xianchao Zhang}, {and} \bibinfo{person}{Xinyue Liu}.} \bibinfo{year}{2021}\natexlab{}.
\newblock \showarticletitle{Spectral embedding network for attributed graph clustering}.
\newblock \bibinfo{journal}{\emph{Neural Networks}}  \bibinfo{volume}{142} (\bibinfo{year}{2021}), \bibinfo{pages}{388--396}.
\newblock


\bibitem[Zhu et~al\mbox{.}(2020)]%
        {GRACE}
\bibfield{author}{\bibinfo{person}{Yanqiao Zhu}, \bibinfo{person}{Yichen Xu}, \bibinfo{person}{Feng Yu}, \bibinfo{person}{Qiang Liu}, \bibinfo{person}{Shu Wu}, {and} \bibinfo{person}{Liang Wang}.} \bibinfo{year}{2020}\natexlab{}.
\newblock \showarticletitle{Deep graph contrastive representation learning}.
\newblock \bibinfo{journal}{\emph{arXiv preprint arXiv:2006.04131}} (\bibinfo{year}{2020}).
\newblock


\end{thebibliography}


\end{document}